\documentclass{article}

\usepackage[preprint,nonatbib]{neurips_2023} 

\usepackage[utf8]{inputenc} 
\usepackage[T1]{fontenc}    
\usepackage{hyperref}       
\usepackage{url}            
\usepackage{booktabs}       
\usepackage{amsfonts}       
\usepackage{nicefrac}       
\usepackage{microtype}      
\usepackage[table]{xcolor}  

\title{Scaling Experiments in Self-Supervised Cross-Table Representation Learning}

\author{%
  Maximilian Schambach${}^{1}$\thanks{Equal contribution.} \\
  \And
  Dominique Paul${}^{1}$\footnotemark[1] \\
  \And
 Johannes S.\ Otterbach${}^{2}$\thanks{Work done while at Merantix Momentum.} \\
 \AND 
  \\[-6mm]${}^{1}$Merantix Momentum, ${}^{2}$ nyonic\\
  Berlin, Germany\\
  \texttt{maximilian.schambach@merantix.com}
}

\renewcommand{\paragraph}[1]{\textbf{#1}\hspace{1em}}
\usepackage{floatrow}

\usepackage{subcaption}
\usepackage{multirow}
\usepackage{rotating}
\usepackage{pdflscape}
\usepackage{tabularx}
\usepackage{wrapfig}
\usepackage[
detect-all,
detect-inline-family=math,
print-unity-mantissa=false,
]{siunitx}
\usepackage{graphicx}
\usepackage[
backend=bibtex,
style=numeric,
maxcitenames=2,
maxbibnames=7,
doi=false,
isbn=false,
url=false,
eprint=false
]{biblatex}
\usepackage[font={normalsize}]{caption}

\begin{document}

\maketitle

\begin{abstract}
To analyze the scaling potential of deep tabular representation learning models, we introduce a novel Transformer-based architecture specifically tailored to tabular data and cross-table representation learning by utilizing table-specific tokenizers and a shared Transformer backbone.
Our training approach encompasses both single-table and cross-table models, trained via missing value imputation through a self-supervised masked cell recovery objective.
To understand the scaling behavior of our method, we train models of varying sizes, ranging from approximately \num{1e4} to \num{1e7} parameters. 
These models are trained on a carefully curated pretraining dataset, consisting of 135\,M training tokens sourced from 76 diverse datasets.
We assess the scaling of our architecture in both single-table and cross-table pretraining setups by evaluating the pretrained models using linear probing on a curated set of benchmark datasets and comparing the results with conventional baselines.
\end{abstract}

\section{Introduction}
\label{sec:introduction}
Tabular data is abundant in many real-world applications across industries as well as research domains and has been argued to be the data type with the highest potential for AI impact \cite{chui_notes_2018}. 
Nevertheless, on tabular data, deep learning approaches fail to consistently outperform established boosting implementations such as XGBoost, LightGBM, and CatBoost \cite{chen_xgboost_2016,ke_lightgbm_2017,dorogush_catboost_2017,grinsztajn_why_2022}.
 Nevertheless, the success of the Transformer architecture~\cite{vaswani_attention_2017} and self-supervised learning applied to large datasets in natural language and computer vision has motivated similar methods in the tabular domain. However, the scaling behavior of these approaches has not been investigated. 
 This is mostly due to the fact that tabular benchmark data is often small and separate models are trained for each table, requiring that the models remain small for fast training and to avoid over-parametrization.
This limits the scaling potential of the underlying architecture as both the model size as well as the training data would need to be scaled for a consistent increase in performance as shown in the language and vision domain~\cite{kaplan_scaling_2020,he_masked_2022}. 
For most tables, however, accessing or creating more data is not possible.

In order to scale tabular deep learning approaches, the architecture needs to be able to generalize across multiple tables so that a large heterogeneous training corpus can be used.
Furthermore, cross-table generalization amortizes the increased costs of training a large versatile model as opposed to training table-specific ones.
Besides a potential performance gain from increased scale, a tabular general-purpose model that generalizes across multiple tables is of practical importance.
For example, pretrained tabular backbones lend themselves as feature extractors and could be of interest in the zero- and few-show regime, with no or only limited training data, as well as in joint representation learning to be used with language models or incorporating inter-table dependencies in relational databases. 

While a variety of cross-table approaches based on Large Language Models (LLMs) have been proposed in the past~\cite{herzig_tapas_2020,yin_tabert_2020,yang_tableformer_2022}, table-specific architectures are extremely scarce.
Despite showing first promising results, we believe the potential of LLMs in the context of tabular data to be limited mainly due to the technical challenges around tokenization which we will discuss in detail.
On the other hand, simple and straightforward Transformer-based architectures in the tabular domain are the exception~\cite{gorishniy_revisiting_2021,zhu_xtab_2023} while the field is scattered with, in our opinion, complex and sometimes convoluted architectures.
We believe a solid understanding of a Transformer-based tabular architecture, and, in particular, the preceding table tokenization, to be the core of future developments and a successful scaling of architectures towards a new state of the art.
In our opinion, the full potential of existing approaches, namely self-supervised Transformer-based architectures, has yet to be understood.

To address these limitations, we propose a clean and simple Transformer-based architecture, similar to the FT-Transformer~\cite{gorishniy_revisiting_2021}, and generalize the architecture for cross-table self-supervised pretraining via masked cell recovery. 
%
Overall, our main contributions are as follows:
\begin{itemize}
    \item We propose a novel architecture and training pipeline for cross-table pretraining based on self-supervised masked cell recovery. This loss can be naturally interpreted as multi-variate value imputation, a formidable problem in real-world applications.
    \item We investigate the scaling behavior of the proposed approach both in a single- as well as a cross-table pretraining setup. We do so by training four model configurations with backbone sizes ranging from roughly \num{1E4} to \num{1E7} parameters using a large curated heterogeneous pretraining corpus of 76 datasets and evaluating the pretrained models via linear probing using a small curated collection of benchmark datasets.
\end{itemize}


\section{Cross-Table Representation Learning}\label{sec:xtable-learning}
While a wide range of approaches has been proposed in the context of learning representations for single tables, covering both supervised~\cite{huang_tabtransformer_2020,gorishniy_revisiting_2021} as well as self-supervised methods~\cite{ucar_subtab_2021,arik_tabnet_2021,somepalli_saint_2022,bahri_scarf_2022,rubachev_revisiting_2022}, how to best design architectures for learning representations across multiple tables is still an open question in the community.
Following the tremendous success of deep learning in natural language and computer vision, Transformer-based architectures trained via self-supervision at scale are most promising to push the state of the art in tabular representation learning and perhaps finally surpass the strong conventional baselines.
However, as opposed to natural language or computer vision, where tokenization and embedding methods naturally generalize across a wide range of datasets, the characteristics of tabular data are table-specific.
Notably, different tables usually have different numbers of columns with numerical and categorical features, as well as column-specific statistics.
That is, even if column names have a similar label indicating a shared semantic, the corresponding (joint and marginalized) statistics may be extremely diverse.
Moreover, unlike language or images, tabular data does not possess a natural ordering and is invariant against column and row permutations.

\paragraph{Cross-table tokenization} 
Tokenization transforms tables (or individual rows) into a sequence of tokens, which are subsequently embedded in a shared embedding space and processed by the model backbone.
In the single-table case, tokenization can be achieved by a combination of conventional tabular encoding and subsequent embedding~\cite{gorishniy_revisiting_2021,gorishniy_embeddings_2022}. 
Numerical features can be tokenized via standardization or quantile transformation while categoricals can be tokenized via integer or one-hot encoding~\cite{james2013introduction}. 
Linear projections or lookup embeddings map the tokens into the embedding space.
However, a cross-table generalization of these approaches is not straightforward and has only recently been proposed within the XTab framework~\cite{zhu_xtab_2023}.
Here, table-specific tokenizers are used to extend the FT-Transformer approach~\cite{gorishniy_revisiting_2021}, whereas the shared backbone contextualizes the embeddings.

A currently popular approach to cross-table tokenization and representation learning is to serialize a table's row into a string, e.g.\ ``\texttt{[Column A]} is \texttt{[Value 1]}, \texttt{[Column B]} is ...''), and then use a pretrained LLM to generate the row's embeddings. 
Many works exist in this area, notably utilizing pretrained BERT models~\cite{herzig_tapas_2020,yin_tabert_2020,trabelsi_strubert_2022,wang_anypredict_2023}, as well as GPT-style generative architectures~\cite{borisov_language_2023,zhang_generative_2023}. \textcite{badaro_transformers_2023,hegselmann_tabllm_2023} discuss and compare multiple forms of table serialization.
This seemingly straightforward concept of table serialization and text-based tokenization comes with a few challenges and pitfalls. 
(i)~Text tokenizers struggle with numerical features, which are typically broken down into multiple tokens by splitting at the decimal point and other subwords in the vocabulary. 
Recent research has shown that this likely leads to subpar performance on numerical tasks such as arithmetic and financial reasoning~\cite{pi_reasoning_2022,yuan_how_2023,wu_bloomberggpt_2023,li_are_2023}. 
While some workarounds, like character-level tokenization for numeric features, have been used~\cite{zhang_generative_2023}, they don't fully address the core issue and introduce additional complexity by requiring a separate decoder architecture. 
(ii)~The coding scheme is not token-efficient, resulting in an excessive amount of tokens per cell. 
As Transformers scale quadratically with the input's length, the excessive representation length of a row requires more computational power than we believe is necessary.
Hence, the number of columns that can be encoded is limited by the context length of the backbone model. 
(iii)~When using causal language modeling, we need to artificially introduce a column order, despite the table's natural column permutation invariance. 
To break this artificial order, any-order learning needs to be enforced, leading to an exponential overhead in column orders that need to be trained, e.g.\ via permutation augmentation. 
On the other hand, in masked language modeling, the masking of individual tokens is not the same as blanking an entire table cell, requiring special treatment of the masking function.

Drawing parallels with text tokenization in natural language processing, we recognize that tokenization is a nuanced, domain-specific problem. 
Tokenizer developments in natural language have significantly enhanced Transformer-based language models by addressing linguistic and engineering challenges~\cite{mielke_between_2021,zouhar_tokenization_2023}. 
In the same way, tokenization for tabular data demands specialized efforts and meticulous experimentation to optimize its utility and compatibility with Transformer architectures. 

\paragraph{Permutation invariance and imputation loss} 
While the embeddings contextualized by a Transformer are inherently permutation invariant, this invariance is typically explicitly broken by introducing positional encodings~\cite{vaswani_attention_2017,dosovitskiy_image_2021}.
Nevertheless, in particular, LLM-based tabular learning architectures use positional encoding and address the problem, if at all, via permutation augmentation~\cite{pietruszka_stable_2022,borisov_deep_2022,zhang_generative_2023}.
Positional encodings are not helpful for tabular data due to their invariance against column permutations. 
Instead, semantic column encodings, e.g.\ via additive column-specific bias embeddings, can be a useful inductive bias to distinguish between different columns~\cite{gorishniy_revisiting_2021,zhu_xtab_2023}. 

A possible solution are bidirectional models, such as BERT~\cite{devlin_bert_2019}, based on masked token recovery losses, akin to a denoising objective. 
Note that this is not a natural loss for a language, which is typically constructed in a sequential manner. 
However, this objective is most natural for table representation learning. 
As table columns have no natural order, and often suffer from missing values, one can interpret masked cell recovery as an imputation of missing values. 
In fact, this allows for a natural generalization to a table-generative model using Markov Random Field sampling~\cite{wang_bert_2019}.

\paragraph{Cross-table pretraining}
In the supervised case, early works treat tables as images and utilize general-purpose vision backbones~\cite{sun_supertml_2019}, whereas recent approaches such as TransTab~\cite{wang_transtab_2022} are limited to tables from similar domains.
In a different line of research, prior-fitted networks were introduced, recasting the problem to approximate Bayesian inference learned over a large synthetic prior, dubbed TabPFN~\cite{hollmann_tabpfn_2023}.
While useful for practitioners and conceptually interesting, TabPFN is limited to small datasets and classification tasks based on purely numeric features and cannot be scaled naively.

Most self-supervised tabular learning approaches are explored in the single-table domain, ranging from autoencoders~\cite{yoon_vime_2020}, contrastive approaches~\cite{ucar_subtab_2021,darabi_contrastive_2021,somepalli_saint_2022,bahri_scarf_2022}, to more recent masked autoencoding objectives~\cite{arik_tabnet_2021,majmundar_met_2022}.
In the cross-table setup, some works deal with self-supervised representation learning for tables with partially (or largely) overlapping columns~\cite{levin_transfer_2023,onishi_tabret_2023}.
We are aware of only one non-LLM-based architecture for unconstrained tabular representation learning, namely the recently proposed XTab framework~\cite{zhu_xtab_2023}.
XTab generalizes the FT-Transformer to multiple tables via table-specific tokenizers and otherwise uses its exact hyperparameter configuration. 
Notably, XTab's Transformer backbone has less than 1\,M trainable parameters.

\begin{figure}
    \centering
    \includegraphics[width=\textwidth]{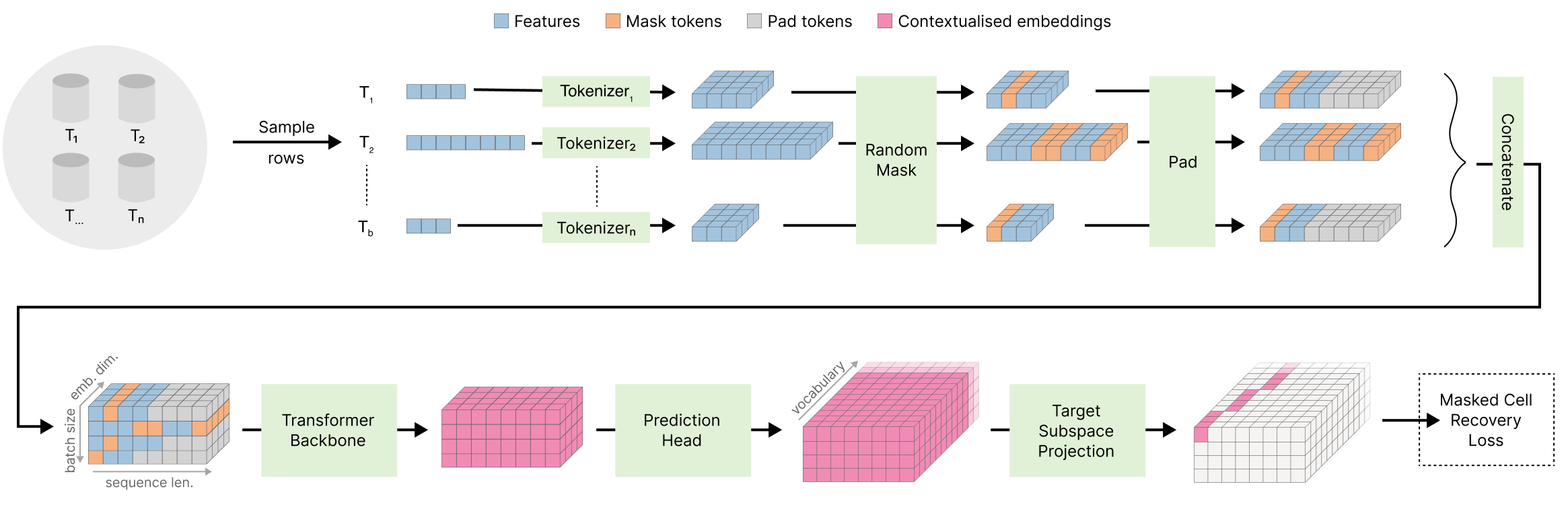}\vspace{-3mm}
    \caption{Overview of the proposed cross-table pertaining architecture. Individual tables are tokenized using table-specific tokenizers, including numerical as well as categorical features, and processed by a shared Transformer backbone.}
    \label{fig:architecture-overview}
\end{figure}

\section{Proposed Approach}\label{sec:proposed-approach}
%
%

We propose a simple Transformer-based architecture and training pipeline for cross-table pretraining that minimizes inductive biases as shown in Figure~\ref{fig:architecture-overview}.
This way, the proposed approach can be used as a baseline for further experimentation, for example around cross-table tokenization techniques.
Our approach builds on the FT-Transformer~\cite{gorishniy_revisiting_2021} and is similar to the recently proposed XTab framework~\cite{zhu_xtab_2023}, with a few important distinctions which we outline in the following.

\paragraph{Tokenization}
We employ table-specific tokenizers and use quantile encoding of numerical features combined with look-up embeddings as opposed to quantile transformation with subsequent linear projection embeddings used in FT-Transformer, XTab, and other approaches~\cite{gorishniy_embeddings_2022}.
That is, instead of transforming the features in order to normalize the column distributions, we encode each value using its quantile index.
Encoding numericals as quasi-categorical values makes the further treatment of all columns uniform.
It simplifies the overall setup and makes the implementation easier to optimize, e.g.\ via vectorization. 
As all values are treated equally, there is no need to distinguish between numericals and categoricals at inference. 
Hence, balancing classification and regression losses is not necessary.
However, the gained flexibility and robustness come at the cost of a quantization error and increased numbers of learnable embedding parameters, depending on the number of quantiles chosen. However, combining a low embedding space and linear up-projection can counter this problem, which we plan to address in future work.
Furthermore, the ordinal character of the encodings is lost without explicit additional treatment.
For categoricals, we use standard integer encoding and embedding via learnable look-up embeddings.
Numerical features with less than 20 unique values are treated directly as categoricals.
Finally, missing values are encoded as an additional NAN category for both numerical as well as categorical features.
Sample statistics needed for the encoding , such as the quantiles, are estimated separately for each dataset using a fixed amount of 10,000 samples each before training.

Finally, we did not add any further additive encodings such as positional encodings and table- or column-specific bias terms, to minimize inductive biases and to retain the permutation invariance of the architecture.
The column- and table-specific characteristics have to be learned by each embedding individually.
While our current work only uses the minimal architectural requirements we see different types of additive encodings as an interesting prospective research question.

\paragraph{Data interleaving}
To obtain rows from multiple tables, we sample from a large hetero\-geneous corpus of tables, which we describe in detail in Section~\ref{sec:experiments-datasets}.
We choose to perform stratified sampling, that is, in every batch the occurrence of each dataset is equally likely, regardless of the dataset size.
This way, we sample uniformly from tasks and domains instead of sampling uniformly from the union of datasets.
As a consequence, smaller datasets are iterated over more often than large ones.
To process these samples in a single batch, we add a learnable padding token to each sequence up to the maximum number of tokens per batch.
This is vastly different from XTab, which utilizes a federated learning approach, deploying the table-specific tokenizers on individual GPUs.
By processing inter-table samples natively, we are able to scale the required hardware independently of the number of tables contained in the pretraining dataset.
In fact, we perform all experiments on a single GPU.
Our approach can easily be further parallelized using standard techniques from distributed training.

\paragraph{Contextualization and learning objective}
The interleaved batch of tokens are contextualized by a single Transformer backbone.
In line with FT-Transformer and XTab, we use the pre-norm variant due to better performance and stability in the natural language context~\cite{wang_learning_2019}.
For self-supervised pretraining, we use the masked cell recovery objective -- the tabular analog of masked language modelling (MLM).
A random subset of tokens per cell is masked with a learnable Mask token and the training objective is to reconstruct the masked values from the contextualized embedding of the corresponding masked token.
We note that this is a natural loss for the case of tabular data, as opposed to MLM. 
Masked tokens can be interpreted as missing values, a common occurrence in practical table modeling problems, and the recovery objective is simply the imputation of its value. 
Compared to traditional imputation methods such as univariate mean, median, or mode estimation, the imputation loss is multi-variate in nature. 
Hence it can capture richer dependencies between columns and other missing values, that are not able to be captured with standard methods.
In the cross-table regime, this loss has been shown to perform better than contrastive pretraining while being more lightweight~\cite{zhu_xtab_2023}.
As opposed to XTab, we fully replace masked tokens with a single learnable mask embedding instead of random values drawn from the marginalized distribution.
We believe this to yield a stronger training signal, but a comparison is left for future works.
Note that, in order to obtain a uniform masking rate for all tasks, masking is performed before padding of the tokens.

For the cell recovery, the contextualized masked tokens are projected by a linear layer into the corresponding target probability space.
As all values have been effectively encoded into categoricals, we optimize for classification via minimization of the cross-entropy loss.
Unlike XTab, we do not use table-specific target heads but perform the target projection into the union of the individual column's target probability spaces.
More precisely, given the individual column-specific target probability spaces $\mathcal{C}_{ij}$ for column $j$ of dataset $i$, the full target probability space is modeled as their direct product, $\mathcal{C} = \prod_{i=1}^{M}\prod_{j=1}^{N_i} \mathcal{C}_{ij}$.
However, the calculation of the cross-entropy for each token is restricted to its individual subspace via binary masking corresponding to an orthogonal projection onto $\mathcal{C}_{ij}$.
\section{Datasets}\label{sec:experiments-datasets}
\setlength{\columnsep}{1.25em}
\begin{wrapfigure}{r}{0.4\textwidth}    
    \vspace{-5mm}
    \centering
    \includegraphics[width=\linewidth]{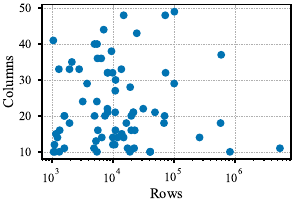}\vspace{2mm}
    \caption{Column and row statistics of the individual datasets contained in our curated pretraining corpus.}\vspace{-6mm}
    \label{fig:training-data}
\end{wrapfigure}
\paragraph{Pretraining corpus}
In order to perform meaningful scaling experiments, sufficient training data is required.
As of now, heterogeneous high-quality tabular training data is not widely available.
Instead, we chose to create a large heterogeneous training corpus by utilizing several tabular benchmark datasets of different tasks and sizes.
For benchmarking, we then restrict ourselves to a small set of curated datasets as discussed next.

We gather datasets from multiple OpenML collections ~\cite{OpenML2013}, and only kept datasets with more than 1000 rows and 10 to 50 columns. We discarded categorical columns that have more than 64 unique values. 
Finally, we manually deduplicated the remaining datasets.

In total, we obtain a corpus containing 76 tables including 30 binary and 26 multiclass classification tasks as well as 20 regression datasets of different widths and sizes.
Overall, the pretraining corpus contains ca.\ 135\,M tokens in total.
Using the previously discussed table-specific tokenization approach, we obtain a token vocabulary size, i.e.\ the number of unique numerical quantiles and categories to be embedded via look-up, of roughly 66\,k.
As a comparison, the BERT language model was trained using a vocabulary size of about 30\,k, whereas GPT-2 used ca.\ 50\,k.
The feature and sample statistics of our pretraining corpus are shown in Figure~\ref{fig:training-data}.
More detailed information on the datasets and statistics are presented in Appendix~\ref{app:dataset-details}.

\paragraph{Benchmark datasets}
Instead of evaluating on a similarly large corpus of datasets or curating a larger set of datasets and splitting it into two folds similar to XTab, we believe a small curated set to be more suitable for investigating these early scaling experiments as opposed to average rank performance across a large benchmark suite. 
This way, we anticipate gaining more nuanced insights into the performance behavior.
For these reasons, we followed the work by \textcite{borisov_deep_2022} and use five tabular datasets for our evaluation, namely HELOC, California Housing, Adult Income, Cover Type, and Higgs, details of which are shown in Appendix~\ref{app:dataset-details}.
These datasets cover a range of tasks (binary and multi-class classification, as well as regression), different numbers and types of columns (from 9 to 55 features), as well as sizes, ranging from roughly 10\,k to 10\,M samples per dataset.
Even in the single-table case, we expect a Transformer-based model to perform severely differently across these five datasets.
We split each dataset into \SI{60}{\percent} used for pretraining and \SI{40}{\percent} evaluated via a 5-fold cross-validation. 
We describe the pretraining and evaluation procedure in detail next.

\section{Experiment Description}
We perform scaling experiments for the proposed architecture using self-supervised pretraining in the single-table as well as the cross-table setup.
In total, we investigate four different model configurations, covering four orders of magnitudes in terms of the backbone model's parameter count, ranging from 13\,k to 16\,M.
Due to limitations with respect to the dataset sizes, for the single-table case, we evaluate models S, M, and L, whereas M, L, and XL are considered in the cross-table case.

\paragraph{Single-table evaluation} Serving as a baseline, we investigate the scaling behavior of our approach in the single-table case. 
That is, for each table in our benchmark suite, we train a separate model via the imputation loss using the mentioned \SI{60}{\percent} pretraining set.
We then evaluate the task-specific performance of the pretrained model via linear probing using 5-fold validation on the remaining \SI{40}{\percent} of each benchmark dataset.
Linear probing is a well-established method to assess the quality of embeddings obtained via self-supervised pretraining and effectively corresponds to learning a linear projection layer supervised on the table-specific task.
Hence, linear probing investigates the linear separability of the table representations with respect to a specific downstream task which the model was not explicitly trained on.
Note that we evaluate the pretraining performance and do not perform any supervised fine-tuning of the tokenizers or backbone, which we leave for future investigations.

\paragraph{Cross-table evaluation} Secondly, we investigate the cross-table case.
Here, each model configuration is pretrained using the imputation loss on the large pretraining corpus.
As our architecture uses table-specific tokenizers, the cross-table pretrained models cannot directly be investigated on the benchmark datasets.
To this end, we again use the table-specific pretraining portion and train the corresponding tokenizers for the pretrained model.
To observe the transferability during training, we checkpoint the pretraining models every 250\,M training tokens and evaluate all checkpointed models via linear probing on all benchmark datasets.
Importantly, for a direct and fair comparison, we also use the same self-supervised learning objective here as in the single-table case to be able to assess the impact of cross-table pretraining.
In this evaluation, we perform two variations: one where the pretrained backbone is frozen and only the tokenizer is trained, and one where the tokenizers and backbone weights are trained jointly.
The obtained models are then evaluated via linear probing in full analogy to the single-table case using 5-fold cross-validation on the remaining portion of each dataset.
Again, we do not perform any supervised fine-tuning.

We want to point out that in both cases a comparison to baselines is challenging, as existing methods, such as boosted trees, are trained in a supervised fashion on a single table. This is in stark contrast to this work, which uses self-supervised training without labeled targets and simply uses the representation features to train a linear model on top to predict the target.
Furthermore, a comparison to other cross-table architectures is difficult, as the only existing approach, XTab, is trained in a federated setup requiring a training cluster of, in our case, 76 GPUs, which is outside our computational budget.

\paragraph{Hyperparameters and training}
Trainings are performed via mini-batch stochastic gradient descent using the AdamW optimizer~\cite{loshchilov_decoupled_2017} with the default parameters.
In the single-table experiments, we choose a batch size of \num{2048} which we reduce to \num{512} for the cross-table pretraining due to memory constraints.
In total, we use 5\,M, 10\,M, and 25\,M samples for pretraining the S, M, and L model in the single-table cases, respectively. 
In the cross-table case, we train all model configurations using 75\,M samples, i.e.\ rows.
The total number of training \emph{tokens} is calculated by summing the number of cells for all samples excluding Padding tokens.
For the learning rate, we choose a warmup phase for the first \SI{10}{\percent} of training samples, linearly increasing the learning rate from \num{5E-5} to \num{1E-3}, and a cosine decay to 0 for the remaining \SI{90}{\percent} of training samples.
We employ a global weight decay, i.e. an $L_2$-norm regularization, of \num{1e-2}.
Throughout, we use a dropout rate of \SI{10}{\percent} during training.
For all experiments, we use a masking fraction of \SI{25}{\percent}.
More details on the used hyperparameters are given in Appendix~\ref{app:technical-details}.
All experiments are conducted using compute nodes with 8 CPU cores, \SI{32}{GB} of RAM, and a single Nvidia L4 GPU.

\paragraph{Baseline methods} For comparison, we evaluate two baseline approaches.
We investigate per-table performance using XGBoost~\cite{chen_xgboost_2016}, as well as a simple linear model using the raw features as predictors.
Naturally, these methods are fitted on each benchmark dataset separately and do not allow for cross-table generalization.
In all cases, we do not perform any hyperparameter optimization -- including our proposed approach.
As we use a different split of the benchmark data, due to the necessity of setting aside a portion for self-supervised pretraining, we cannot directly compare with the many baselines presented in the paper by \textcite{borisov_deep_2022}.
However, we do not expect the results to be fundamentally different on the splits used here as we follow the identical evaluation protocol via five-fold cross-validation.

\begin{figure}
    \centering
    \includegraphics[width=\linewidth]{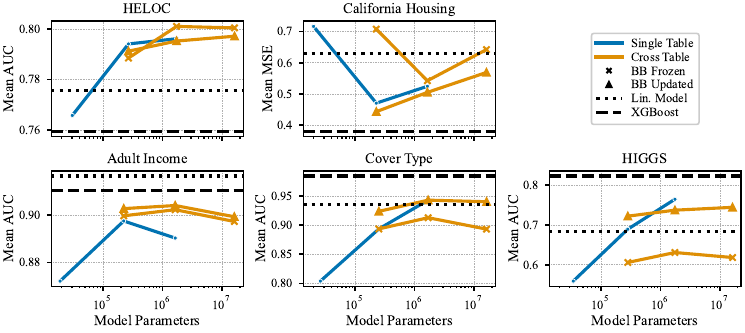}\vspace{-1mm}
    \caption{Mean 5-fold cross-validation linear probing results for the considered benchmark datasets in the case of single-table as well as cross-table pretraining with frozen and updated backbones (BB). The (supervised) performance of a linear model as well as XGBoost are shown for comparison.}
    \label{fig:results-scaling}
\end{figure}

\section{Results}
\label{sec:results}
Our main results investigate the scaling behavior of the different models in terms of their linear probing performance on the benchmark datasets and are shown in Figure~\ref{fig:results-scaling}.

\paragraph{Single-table performance}
Investigating the single-table case, we make the following observations:
First, the imputation objective of recovering masked cell values is indeed informative on the dataset-specific downstream task.
Recall that we do not perform any supervised fine-tuning. 
It indicates that the models are indeed learning multi-variate dependencies to efficiently recover missing values.
That is, despite the model not being trained on the task specifics, the obtained contextualized features show good linear separability with respect to the downstream tasks.
In most cases, in particular for HELOC and HIGGS, the contextualized features have more predictive power than the unprocessed ones as shown by the comparison with the linear model.
Generally, the results are sub-par compared to a non-optimized XGBoost, which, however, is trained in a supervised fashion.
With respect to the backbone size, we see slight improvements with scale:
The linear probing performance increases with the amount of backbone parameters, as expected.
We do, however, observe that in most cases, in particular with smaller datasets, this increase is saturated already with the Medium model configuration while larger datasets, such as Cover Type and HIGGS, do not show this saturation.
This is, to some extent, expected as the amount of training data has to be scaled with increasing backbone parameter counts.
This supports our claimed need for cross-table approaches in order to be able to scale tabular models towards a much larger scale.
The training loss and imputation accuracies for all trained models are provided in Appendix~\ref{app:additional-results}.

\paragraph{Cross-table performance}
Generally, we observe a slight increase in performance when using cross-table pretraining, in particular notable in the HELOC and Adult Income datasets.
Typically, the updating of the backbone parameters jointly with the training of the tokenizer, again in a self-supervised fashion, tends to perform better than the frozen weights obtained during pretraining, with the exception of the HELOC dataset.
Overall, we do not see a strong increase in performance with scale, which indicates that we might be far from optimal dataset sizes to saturate the models and learn meaningful cross-table contextualization patterns within the backbone. 
On the other hand, we also observe that scaling does not hurt performance, which could indicate that increasing the dataset sizes can lead to improvements.
Slight increases can be observed in the HELOC dataset, whereas increased scale actually leads to worse performance in some instances such as the California Housing dataset.
Moreover, we see an interestingly steep increase in imputation accuracy during transfer learning on the benchmark datasets, as shown in Figure~\ref{fig:results-imputation-transfer} in the case of the Adult Income dataset and in Appendix~\ref{app:additional-results} for the remaining ones.
This encourages the usage of the proposed cross-table pretrained model as a multivariate imputation system.

Further, looking more closely at the linear probing performance at several stages during pretraining, which are shown in Appendix~\ref{app:additional-results}, we do not see systematic improvements with longer pretraining.
This is surprising and suggests that the backbone feature processing does not increase in performance with increased pretraining performance.
That is, while we see an increase in pretraining imputation accuracy, this does not directly transfer to improvements with respect to the linear separability of the benchmark tasks, unlike our observations in the single-table case.
This is an interesting observation that could be caused by a number of reasons opening several future research directions.
First, we note that the cross-table pretraining was limited by our compute budget and that all models, in particular the L and XL variants, show further potential in training as shown in Figure~\ref{fig:results-xt-training-loss}.
Here, the training loss of the XL model is hardly saturated and we expect further gains with longer training.
This is less the case in the single-table training, for which we present the loss curves in Appendix~\ref{app:additional-results}, which are limited by the individual dataset sizes and saturate much earlier.
Second, the approach to using table-specific tokenizers comes at the cost of a comparably large parameter overhead.
As previously mentioned, our cross-table pretraining vocabulary contains 66\,k tokens and look-up embeddings, resulting in a large number of additional training parameters as detailed in Appendix~\ref{app:technical-details}.
For comparison, GPT-2 uses a 60\,k subword vocabulary at a size of 1.5\,B parameters and 40\,B training tokens, which is orders of magnitudes larger than the ones used here.
This imbalance of tokenization and backbone parameters could be a reason for the observed behavior.
Continued scaling experimentation is required, while keeping the vocabulary size constant, e.g.\ by using larger pretraining datasets or improving the tokenization efficiency by using a lower-dimensional embedding space combined with a shared upsampling layer.
Finally, we do not investigate supervised fine-tuning here.
For one, it would be interesting to observe whether pretraining boosts supervised fine-tuning, similar to the results obtained in the XTab framework~\cite{zhu_xtab_2023}.
Furthermore, using a supervised objective, either in addition to the self-supervised pretraining or for the benchmark dataset transfer, would allow for introducing a learnable CLS token to aggregate the contextualized embeddings in an adaptive way.
Currently, our evaluation protocol uses mean pooling across the contextualized row tokens, excluding Pad tokens, for linear probing.
This aggregation might smooth out representations with higher predictive performance and is not task-adaptive.
However, in the fully self-supervised case, it is not directly possible to introduce global contextualized representations, e.g.\ via a learnable CLS token.

\paragraph{Limitations and future work}
Our current approach offers several limitations, the most technical of which we previously discussed.
In addition, our current evaluation protocol is limited in scope.
A comparison across more benchmark datasets as well as supervised and unsupervised baselines, such as boosting or LLM-based approaches, is of interest and we plan to address this in the future.
Also, performing a hyperparameter optimization should yield better results for both the considered baselines and the proposed approach, e.g.\ investigating the dropout and masking ratios in detail.
Furthermore, we plan to investigate the cross-table tokenization in detail in future works, for example, the impact of row and table encodings as well as the explicit use of the individual table schemas, for example by using a separate learnable schema embedder.
Finally, we argue there is a great need for more elaborate tabular training data in order to scale tabular models towards model sizes comparable to, e.g., GPT-2 as a first step.
Similarly, benchmarks tailored to the usage of deep learning models need to be further developed and refined.

\begin{figure}
	\begin{floatrow}
		\ffigbox[0.48\textwidth]
		{\captionof{figure}{Imputation accuracy during transfer learning of the cross-table pretrained backbone onto the Adult Income dataset.}%
		\label{fig:results-imputation-transfer}}
		{\includegraphics[width=\linewidth]{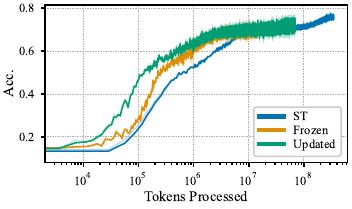}}
		\ffigbox[0.48\textwidth]
		{\captionof{figure}{Pretraining loss curves and imputation accuracy of the cross-table models.}%
		\label{fig:results-xt-training-loss}}
		{\vspace{-2mm}\includegraphics{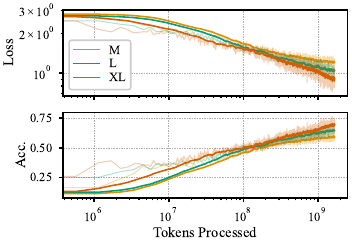}}
	\end{floatrow}%
\end{figure}

\section{Conclusion}
\label{sec:conclusion}
We have presented a novel architecture and training pipeline for cross-table pretraining and conducted scaling experiments that showed first interesting results.
Generally, we see an increase in the linear probing accuracy across several benchmark datasets with larger model scales in both the single- and the cross-table case.
Whereas models trained in a single-table fashion saturated, we saw slight improvements using cross-table pretraining, which was however limited likely due to a lack of training data or compute resources.
We have discussed multiple possible reasons for the observed behavior and interesting further research directions.

{\small\printbibliography}

@misc{zouhar_tokenization_2023,
	title = {Tokenization and the {Noiseless} {Channel}},
	url = {http://arxiv.org/abs/2306.16842},
	doi = {10.48550/arXiv.2306.16842},
	urldate = {2023-09-26},
	publisher = {arXiv},
	author = {Zouhar, Vilém and Meister, Clara and Gastaldi, Juan Luis and Du, Li and Sachan, Mrinmaya and Cotterell, Ryan},
	year = {2023},
	note = {arXiv:2306.16842},
}

@misc{zhang_generative_2023,
	title = {Generative {Table} {Pre}-training {Empowers} {Models} for {Tabular} {Prediction}},
	url = {http://arxiv.org/abs/2305.09696},
	doi = {10.48550/arXiv.2305.09696},
	urldate = {2023-09-07},
	publisher = {arXiv},
	author = {Zhang, Tianping and Wang, Shaowen and Yan, Shuicheng and Li, Jian and Liu, Qian},
	year = {2023},
	note = {arXiv:2305.09696},
}

@misc{yuan_how_2023,
	title = {How {Well} {Do} {Large} {Language} {Models} {Perform} in {Arithmetic} {Tasks}?},
	url = {http://arxiv.org/abs/2304.02015},
	urldate = {2023-09-05},
	publisher = {arXiv},
	author = {Yuan, Zheng and Yuan, Hongyi and Tan, Chuanqi and Wang, Wei and Huang, Songfang},
	year = {2023},
	note = {arXiv:2304.02015},
}

@inproceedings{yoon_vime_2020,
	title = {{VIME}: {Extending} the {Success} of {Self}- and {Semi}-supervised {Learning} to {Tabular} {Domain}},
	volume = {33},
	shorttitle = {{VIME}},
	url = {https://proceedings.neurips.cc/paper/2020/hash/7d97667a3e056acab9aaf653807b4a03-Abstract.html},
	urldate = {2023-09-05},
	booktitle = {Advances in {Neural} {Information} {Processing} {Systems}},
	author = {Yoon, Jinsung and Zhang, Yao and Jordon, James and van der Schaar, Mihaela},
	year = {2020},
	pages = {11033--11043},
}

@article{yang_tableformer_2022,
	title = {{TableFormer}: {Robust} {Transformer} {Modeling} for {Table}-{Text} {Encoding}},
	doi = {10.18653/v1/2022.acl-long.40},
	journal = {Annual Meeting of the Association for Computational Linguistics},
	author = {Yang, Jingfeng and Gupta, Aditya and Upadhyay, Shyam and He, Luheng and Goel, Rahul and Paul, Shachi},
	year = {2022},
	pages = {528--537},
}

@misc{wu_bloomberggpt_2023,
	title = {{BloombergGPT}: {A} {Large} {Language} {Model} for {Finance}},
	shorttitle = {{BloombergGPT}},
	url = {http://arxiv.org/abs/2303.17564},
	doi = {10.48550/arXiv.2303.17564},
	urldate = {2023-09-16},
	publisher = {arXiv},
	author = {Wu, Shijie and Irsoy, Ozan and Lu, Steven and Dabravolski, Vadim and Dredze, Mark and Gehrmann, Sebastian and Kambadur, Prabhanjan and Rosenberg, David and Mann, Gideon},
	year = {2023},
	note = {arXiv:2303.17564},
}

@misc{wang_anypredict_2023,
	title = {{AnyPredict}: {Foundation} {Model} for {Tabular} {Prediction}},
	shorttitle = {{AnyPredict}},
	url = {http://arxiv.org/abs/2305.12081},
	doi = {10.48550/arXiv.2305.12081},
	urldate = {2023-09-07},
	publisher = {arXiv},
	author = {Wang, Zifeng and Gao, Chufan and Xiao, Cao and Sun, Jimeng},
	year = {2023},
	note = {arXiv:2305.12081},
}

@inproceedings{wang_learning_2019,
	title = {Learning {Deep} {Transformer} {Models} for {Machine} {Translation}},
	url = {https://aclanthology.org/P19-1176},
	doi = {10.18653/v1/P19-1176},
	urldate = {2023-09-16},
	booktitle = {Annual {Meeting} of the {Association} for {Computational} {Linguistics}},
	author = {Wang, Qiang and Li, Bei and Xiao, Tong and Zhu, Jingbo and Li, Changliang and Wong, Derek F. and Chao, Lidia S.},
	year = {2019},
	pages = {1810--1822},
}

@inproceedings{wang_bert_2019,
	title = {{BERT} has a {Mouth}, and {It} {Must} {Speak}: {BERT} as a {Markov} {Random} {Field} {Language} {Model}},
	shorttitle = {{BERT} has a {Mouth}, and {It} {Must} {Speak}},
	url = {https://aclanthology.org/W19-2304},
	doi = {10.18653/v1/W19-2304},
	urldate = {2023-09-26},
	booktitle = {{NAACL} {Workshop} on {Methods} for {Optimizing} and {Evaluating} {Neural} {Language} {Generation}},
	author = {Wang, Alex and Cho, Kyunghyun},
	year = {2019},
	pages = {30--36},
}

@misc{ucar_subtab_2021,
	title = {{SubTab}: {Subsetting} {Features} of {Tabular} {Data} for {Self}-{Supervised} {Representation} {Learning}},
	url = {https://app.readcube.com/library/12dc685d-b246-4fde-a409-59b857cce495/item/9258fedb-845d-47cb-b74c-78fe5099b8af},
	doi = {10.48550/arxiv.2110.04361},
	publisher = {arxiv},
	author = {Ucar, Talip and Hajiramezanali, Ehsan and Edwards, Lindsay},
	year = {2021},
	note = {arxiv:2110.04361},
}

@misc{rubachev_revisiting_2022,
	title = {Revisiting {Pretraining} {Objectives} for {Tabular} {Deep} {Learning}},
	doi = {10.48550/arxiv.2207.03208},
	publisher = {arxiv},
	author = {Rubachev, Ivan and Alekberov, Artem and Gorishniy, Yury and Babenko, Artem},
	year = {2022},
	note = {arxiv.2207.03208},
}

@misc{power_grokking_2022,
	title = {Grokking: {Generalization} {Beyond} {Overfitting} on {Small} {Algorithmic} {Datasets}},
	shorttitle = {Grokking},
	url = {http://arxiv.org/abs/2201.02177},
	doi = {10.48550/arXiv.2201.02177},
	urldate = {2023-09-28},
	publisher = {arXiv},
	author = {Power, Alethea and Burda, Yuri and Edwards, Harri and Babuschkin, Igor and Misra, Vedant},
	year = {2022},
	note = {arXiv:2201.02177},
}

@article{nakkiran_deep_2021,
	title = {Deep {Double} {Descent}: {Where} {Bigger} {Models} and {More} {Data} {Hurt}},
	volume = {2021},
	issn = {1742-5468},
	shorttitle = {Deep double descent},
	url = {https://dx.doi.org/10.1088/1742-5468/ac3a74},
	doi = {10.1088/1742-5468/ac3a74},
	number = {12},
	urldate = {2023-09-28},
	journal = {Journal of Statistical Mechanics: Theory and Experiment},
	author = {Nakkiran, Preetum and Kaplun, Gal and Bansal, Yamini and Yang, Tristan and Barak, Boaz and Sutskever, Ilya},
	year = {2021},
	note = {Publisher: IOP Publishing and SISSA},
	pages = {124003},
}

@misc{mielke_between_2021,
	title = {Between words and characters: {A} {Brief} {History} of {Open}-{Vocabulary} {Modeling} and {Tokenization} in {NLP}},
	shorttitle = {Between words and characters},
	url = {http://arxiv.org/abs/2112.10508},
	doi = {10.48550/arXiv.2112.10508},
	urldate = {2023-09-16},
	publisher = {arXiv},
	author = {Mielke, Sabrina J. and Alyafeai, Zaid and Salesky, Elizabeth and Raffel, Colin and Dey, Manan and Gallé, Matthias and Raja, Arun and Si, Chenglei and Lee, Wilson Y. and Sagot, Benoît and Tan, Samson},
	month = dec,
	year = {2021},
	note = {arXiv:2112.10508},
}

@misc{li_are_2023,
	title = {Are {ChatGPT} and {GPT}-4 {General}-{Purpose} {Solvers} for {Financial} {Text} {Analytics}? {An} {Examination} on {Several} {Typical} {Tasks}},
	shorttitle = {Are {ChatGPT} and {GPT}-4 {General}-{Purpose} {Solvers} for {Financial} {Text} {Analytics}?},
	url = {http://arxiv.org/abs/2305.05862},
	doi = {10.48550/arXiv.2305.05862},
	urldate = {2023-09-16},
	publisher = {arXiv},
	author = {Li, Xianzhi and Zhu, Xiaodan and Ma, Zhiqiang and Liu, Xiaomo and Shah, Sameena},
	month = may,
	year = {2023},
	note = {arXiv:2305.05862},
}

@book{james2013introduction,
	title = {An {Introduction} to {Statistical} {Learning}},
	volume = {112},
	publisher = {Springer},
	author = {James, Gareth and Witten, Daniela and Hastie, Trevor and Tibshirani, Robert and {others}},
	year = {2013},
}

@misc{kaplan_scaling_2020,
	title = {Scaling {Laws} for {Neural} {Language} {Models}},
	url = {https://app.readcube.com/library/12dc685d-b246-4fde-a409-59b857cce495/item/70d30dc0-adf9-451a-b044-e70f7952f3ef},
	doi = {10.48550/arxiv.2001.08361},
	publisher = {arxiv},
	author = {Kaplan, Jared and McCandlish, Sam and Henighan, Tom and Brown, Tom B and Chess, Benjamin and Child, Rewon and Gray, Scott and Radford, Alec and Wu, Jeffrey and Amodei, Dario},
	year = {2020},
	note = {arxiv:2001.08361},
}

@misc{huang_tabtransformer_2020,
	title = {{TabTransformer}: {Tabular} {Data} {Modeling} {Using} {Contextual} {Embeddings}},
	doi = {10.48550/arxiv.2012.06678},
	publisher = {arxiv},
	author = {Huang, Xin and Khetan, Ashish and Cvitkovic, Milan and Karnin, Zohar},
	year = {2020},
	note = {arxiv:2012.06678},
}

@article{herzig_tapas_2020,
	title = {{TAPAS}: {Weakly} {Supervised} {Table} {Parsing} via {Pre}-training},
	url = {https://doi.org/10.18653/v1/2020.acl-main.398},
	doi = {10.48550/arxiv.2004.02349},
	journal = {Annual Meeting of the Association for Computational Linguistics},
	author = {Herzig, Jonathan and Nowak, Paweł Krzysztof and Müller, Thomas and Piccinno, Francesco and Eisenschlos, Julian Martin},
	year = {2020},
}

@article{grinsztajn_why_2022,
	title = {Why {Do} {Tree}-{Based} {Models} {Still} {Outperform} {Deep} {Learning} on {Typical} {Tabular} {Data}?},
	url = {https://proceedings.neurips.cc/paper_files/paper/2022/file/0378c7692da36807bdec87ab043cdadc-Paper-Datasets_and_Benchmarks.pdf},
	journal = {Advances in Neural Information Processing Systems},
	author = {Grinsztajn, Leo and Oyallon, Edouard and Varoquaux, Gael},
	year = {2022},
	pages = {507--520},
}

@inproceedings{dorogush_catboost_2017,
	title = {{CatBoost}: {Gradient} {Boosting} with {Categorical} {Features} {Support}},
	shorttitle = {{CatBoost}},
	url = {http://arxiv.org/abs/1810.11363},
	doi = {10.48550/arXiv.1810.11363},
	urldate = {2023-09-05},
	booktitle = {{NeurIPS} {ML} {Systems} {Workshop}},
	author = {Dorogush, Anna Veronika and Ershov, Vasily and Gulin, Andrey},
	year = {2017},
}

@article{devlin_bert_2019,
	title = {{BERT}: {Pre}-training of {Deep} {Bidirectional} {Transformers} for {Language} {Understanding}},
	url = {https://app.readcube.com/library/12dc685d-b246-4fde-a409-59b857cce495/item/acbb6fd0-f6cf-4140-8d5c-56e562c8bd66},
	doi = {10.48550/arxiv.1810.04805},
	journal = {North American Chapter of the Association for Computational Linguistics: Human Language Technologies},
	author = {Devlin, Jacob and Chang, Ming-Wei and Lee, Kenton and Toutanova, Kristina},
	year = {2019},
}

@techreport{chui_notes_2018,
	title = {Notes {From} the {AI} {Frontier}: {Insights} from {Hundreds} of {Use} {Cases}},
	shorttitle = {Notes from the {AI} frontier},
	institution = {McKinsey Global Institute},
	author = {Chui, Michael and Manyika, James and Miremadi, Mehdi and Henke, Nicolaus and Chung, Rita and Nel, Pieter and Malhotra, Sankalp},
	year = {2018},
}

@inproceedings{chen_xgboost_2016,
	title = {{XGBoost}: {A} {Scalable} {Tree} {Boosting} {System}},
	isbn = {978-1-4503-4232-2},
	shorttitle = {{XGBoost}},
	url = {https://dl.acm.org/doi/10.1145/2939672.2939785},
	doi = {10.1145/2939672.2939785},
	urldate = {2023-09-12},
	booktitle = {{ACM} {SIGKDD} {International} {Conference} on {Knowledge} {Discovery} and {Data} {Mining}},
	publisher = {Association for Computing Machinery},
	author = {Chen, Tianqi and Guestrin, Carlos},
	month = aug,
	year = {2016},
	pages = {785--794},
}

@misc{belkin_fit_2021,
	title = {Fit {Without} {Fear}: {Remarkable} {Mathematical} {Phenomena} of {Deep} {Learning} {Through} the {Prism} of {Interpolation}},
	shorttitle = {Fit without fear},
	url = {http://arxiv.org/abs/2105.14368},
	doi = {10.48550/arXiv.2105.14368},
	urldate = {2023-09-28},
	publisher = {arXiv},
	author = {Belkin, Mikhail},
	month = may,
	year = {2021},
}

@article{wang_transtab_2022,
	title = {{TransTab}: {Learning} {Transferable} {Tabular} {Transformers} {Across} {Tables}},
	volume = {35},
	shorttitle = {{TransTab}},
	url = {https://proceedings.neurips.cc/paper_files/paper/2022/hash/1377f76686d56439a2bd7a91859972f5-Abstract-Conference.html},
	urldate = {2023-09-05},
	journal = {Advances in Neural Information Processing Systems},
	author = {Wang, Zifeng and Sun, Jimeng},
	year = {2022},
	pages = {2902--2915},
}

@inproceedings{onishi_tabret_2023,
	title = {{TabRet}: {Pre}-training {Transformer}-based {Tabular} {Models} for {Unseen} {Columns}},
	shorttitle = {{TabRet}},
	url = {http://arxiv.org/abs/2303.15747},
	urldate = {2023-09-07},
	booktitle = {{ICLR} {Workshop} on {Understanding} {Foundation} {Models}},
	publisher = {arXiv},
	author = {Onishi, Soma and Oono, Kenta and Hayashi, Kohei},
	year = {2023},
	note = {arXiv:2303.15747 [cs]},
}

@inproceedings{he_masked_2022,
	title = {Masked {Autoencoders} {Are} {Scalable} {Vision} {Learners}},
	url = {https://openaccess.thecvf.com/content/CVPR2022/html/He_Masked_Autoencoders_Are_Scalable_Vision_Learners_CVPR_2022_paper.html},
	urldate = {2023-09-13},
	booktitle = {Conference on {Computer} {Vision} and {Pattern} {Recognition}},
	author = {He, Kaiming and Chen, Xinlei and Xie, Saining and Li, Yanghao and Dollár, Piotr and Girshick, Ross},
	year = {2022},
	pages = {16000--16009},
}

@article{darabi_contrastive_2021,
	title = {Contrastive {Mixup}: {Self}- and {Semi}-{Supervised} learning for {Tabular} {Domain}},
	volume = {2108.12296},
	url = {https://app.readcube.com/library/12dc685d-b246-4fde-a409-59b857cce495/item/e2de02db-1ec6-4386-9287-5d1c4e8da6c4},
	doi = {10.48550/arxiv.2108.12296},
	journal = {ArXiv},
	author = {Darabi, Sajad and Fazeli, Shayan and Pazoki, Ali and Sankararaman, Sriram and Sarrafzadeh, Majid},
	year = {2021},
}

@article{vaswani_attention_2017,
	title = {Attention {Is} {All} {You} {Need}},
	url = {https://proceedings.neurips.cc/paper/2017/hash/3f5ee243547dee91fbd053c1c4a845aa-Abstract.html},
	journal = {Advances in Neural Information Processing System},
	author = {Vaswani, Ashish and Shazeer, Noam and Parmar, Niki and Uszkoreit, Jakob and Jones, Llion and Gomez, Aidan N and Kaiser, Lukasz and Polosukhin, Illia},
	year = {2017},
}

@inproceedings{sun_supertml_2019,
	title = {{SuperTML}: {Two}-{Dimensional} {Word} {Embedding} for the {Precognition} on {Structured} {Tabular} {Data}},
	shorttitle = {{SuperTML}},
	url = {https://openaccess.thecvf.com/content_CVPRW_2019/html/Precognition/Sun_SuperTML_Two-Dimensional_Word_Embedding_for_the_Precognition_on_Structured_Tabular_CVPRW_2019_paper.html},
	urldate = {2023-09-07},
	booktitle = {Conference on {Computer} {Vision} and {Pattern} {Recognition} {Workshops}},
	author = {Sun, Baohua and Yang, Lin and Zhang, Wenhan and Lin, Michael and Dong, Patrick and Young, Charles and Dong, Jason},
	year = {2019},
}

@inproceedings{ke_lightgbm_2017,
	title = {{LightGBM}: {A} {Highly} {Efficient} {Gradient} {Boosting} {Decision} {Tree}},
	shorttitle = {{LightGBM}},
	url = {https://papers.nips.cc/paper_files/paper/2017/hash/6449f44a102fde848669bdd9eb6b76fa-Abstract.html},
	urldate = {2023-09-12},
	booktitle = {Advances in {Neural} {Information} {Processing} {Systems}},
	author = {Ke, Guolin and Meng, Qi and Finley, Thomas and Wang, Taifeng and Chen, Wei and Ma, Weidong and Ye, Qiwei and Liu, Tie-Yan},
	year = {2017},
}

@inproceedings{borisov_language_2023,
	title = {Language {Models} are {Realistic} {Tabular} {Data} {Generators}},
	url = {http://arxiv.org/abs/2210.06280},
	doi = {10.48550/arXiv.2210.06280},
	urldate = {2023-09-07},
	booktitle = {International {Conference} on {Learning} {Representations}},
	author = {Borisov, Vadim and Seßler, Kathrin and Leemann, Tobias and Pawelczyk, Martin and Kasneci, Gjergji},
	year = {2023},
}

@article{borisov_deep_2022,
	title = {Deep {Neural} {Networks} and {Tabular} {Data}: {A} {Survey}},
	issn = {2162-237X},
	url = {https://app.readcube.com/library/12dc685d-b246-4fde-a409-59b857cce495/item/9993f25a-3f5b-4312-89c6-d719b4732ee8},
	doi = {10.1109/tnnls.2022.3229161},
	number = {99},
	journal = {IEEE Transactions on Neural Networks and Learning Systems},
	author = {Borisov, Vadim and Leemann, Tobias and Sessler, Kathrin and Haug, Johannes and Pawelczyk, Martin and Kasneci, Gjergji},
	year = {2022},
	pages = {1--21},
}

@article{badaro_transformers_2023,
	title = {Transformers for {Tabular} {Data} {Representation}: {A} {Survey} of {Models} and {Applications}},
	volume = {11},
	shorttitle = {Transformers for {Tabular} {Data} {Representation}},
	url = {https://aclanthology.org/2023.tacl-1.14},
	doi = {10.1162/tacl_a_00544},
	urldate = {2023-09-07},
	journal = {Transactions of the Association for Computational Linguistics},
	author = {Badaro, Gilbert and Saeed, Mohammed and Papotti, Paolo},
	year = {2023},
	pages = {227--249},
}

@inproceedings{hollmann_tabpfn_2023,
	title = {{TabPFN}: {A} {Transformer} {That} {Solves} {Small} {Tabular} {Classification} {Problems} in a {Second}},
	url = {https://openreview.net/forum?id=cp5PvcI6w8_},
	doi = {10.48550/arxiv.2207.01848},
	booktitle = {International {Conference} on {Learning} {Representations}},
	author = {Hollmann, Noah and Müller, Samuel and Eggensperger, Katharina and Hutter, Frank},
	year = {2023},
}

@inproceedings{pi_reasoning_2022,
	address = {Abu Dhabi, United Arab Emirates},
	title = {Reasoning {Like} {Program} {Executors}},
	url = {https://aclanthology.org/2022.emnlp-main.48},
	doi = {10.18653/v1/2022.emnlp-main.48},
	urldate = {2023-09-11},
	booktitle = {Proceedings of the 2022 {Conference} on {Empirical} {Methods} in {Natural} {Language} {Processing}},
	publisher = {Association for Computational Linguistics},
	author = {Pi, Xinyu and Liu, Qian and Chen, Bei and Ziyadi, Morteza and Lin, Zeqi and Fu, Qiang and Gao, Yan and Lou, Jian-Guang and Chen, Weizhu},
	month = dec,
	year = {2022},
	pages = {761--779},
}

@article{majmundar_met_2022,
	title = {{MET}: {Masked} {Encoding} for {Tabular} {Data}},
	url = {https://openreview.net/forum?id=vMHs3HR7r0A},
	doi = {10.48550/arxiv.2206.08564},
	journal = {NeurIPS Table Representation Learning Workshop},
	author = {Majmundar, Kushal and Goyal, Sachin and Netrapalli, Praneeth and Jain, Prateek},
	year = {2022},
}

@article{gorishniy_revisiting_2021,
	title = {Revisiting {Deep} {Learning} {Models} for {Tabular} {Data}},
	url = {https://app.readcube.com/library/12dc685d-b246-4fde-a409-59b857cce495/item/d581b938-44d2-4d7f-8456-50e4c12c8572},
	doi = {10.48550/arxiv.2106.11959},
	journal = {Advances in Neural Information Processing Systems},
	author = {Gorishniy, Yury and Rubachev, Ivan and Khrulkov, Valentin and Babenko, Artem},
	year = {2021},
}

@article{zhu_xtab_2023,
	title = {{XTab}: {Cross}-table {Pretraining} for {Tabular} {Transformers}},
	url = {https://app.readcube.com/library/12dc685d-b246-4fde-a409-59b857cce495/item/2d350e56-e933-4948-a970-e3d14fca505d},
	doi = {10.48550/arxiv.2305.06090},
	journal = {International Conference On Machine Learning},
	author = {Zhu, Bingzhao and Shi, Xingjian and Erickson, Nick and Li, Mu and Karypis, George and Shoaran, Mahsa},
	year = {2023},
}

@article{hegselmann_tabllm_2023,
	title = {{TabLLM}: {Few}-shot {Classification} of {Tabular} {Data} with {Large} {Language} {Models}},
	url = {https://proceedings.mlr.press/v206/hegselmann23a.html},
	doi = {10.48550/arxiv.2210.10723},
	journal = {International Conference on Artificial Intelligence and Statistics},
	author = {Hegselmann, Stefan and Buendia, Alejandro and Lang, Hunter and Agrawal, Monica and Jiang, Xiaoyi and Sontag, David},
	year = {2023},
}

@article{arik_tabnet_2021,
	title = {{TabNet}: {Attentive} {Interpretable} {Tabular} {Learning}},
	volume = {35},
	issn = {2159-5399},
	url = {https://app.readcube.com/library/12dc685d-b246-4fde-a409-59b857cce495/item/42f9113d-ecd4-4430-a9da-c3607fa0b725},
	doi = {10.1609/aaai.v35i8.16826},
	number = {8},
	journal = {AAAI Conference on Artificial Intelligence},
	author = {Arik, Sercan Ö and Pfister, Tomas},
	year = {2021},
	pages = {6679--6687},
}

@article{gorishniy_embeddings_2022,
	title = {On {Embeddings} for {Numerical} {Features} in {Tabular} {Deep} {Learning}},
	url = {https://proceedings.neurips.cc/paper_files/paper/2022/hash/9e9f0ffc3d836836ca96cbf8fe14b105-Abstract-Conference.html},
	doi = {10.48550/arxiv.2203.05556},
	journal = {Advances in Neural Information Processing Systems},
	author = {Gorishniy, Yury and Rubachev, Ivan and Babenko, Artem},
	year = {2022},
}

@article{dosovitskiy_image_2021,
	title = {An {Image} is worth 16x16 words: {Transformers} for image recognition at scale},
	url = {https://openreview.net/forum?id=YicbFdNTTy},
	journal = {International Conference on Learning Representations},
	author = {Dosovitskiy, Alexey and Beyer, Lucas and Kolesnikov, Alexander and Weissenborn, Dirk and Zhai, Xiaohua and Unterthiner, Thomas and Dehghani, Mostafa and Minderer, Matthias and Heigold, Georg and Gelly, Sylvain and Uszkoreit, Jakob and Houlsby, Neil},
	year = {2021},
}

@article{bahri_scarf_2022,
	title = {{SCARF}: {Self}-{Supervised} {Contrastive} {Learning} using {Random} {Feature} {Corruption}},
	url = {https://openreview.net/forum?id=CuV_qYkmKb3},
	doi = {10.48550/arxiv.2106.15147},
	journal = {International Conference on Learning Representations},
	author = {Bahri, Dara and Jiang, Heinrich and Tay, Yi and Metzler, Donald},
	year = {2022},
}

@article{trabelsi_strubert_2022,
	title = {{StruBERT}: {Structure}-aware {BERT} for {Table} {Search} and {Matching}},
	url = {https://app.readcube.com/library/12dc685d-b246-4fde-a409-59b857cce495/item/a263e1fd-0c64-4d06-840a-5ff4096f3cc5},
	doi = {10.1145/3485447.3511972},
	journal = {ACM Web Conference},
	author = {Trabelsi, Mohamed and Chen, Zhiyu and Zhang, Shuo and Davison, Brian D and Heflin, Jeff},
	year = {2022},
	pages = {442--451},
}

@article{yin_tabert_2020,
	title = {{TaBERT}: {Pretraining} for {Joint} {Understanding} of {Textual} and {Tabular} {Data}},
	url = {https://app.readcube.com/library/12dc685d-b246-4fde-a409-59b857cce495/item/900ed783-bc7d-42f3-b588-0fd236ef214b},
	doi = {10.18653/v1/2020.acl-main.745},
	journal = {Annual Meeting of the Association for Computational Linguistics},
	author = {Yin, Pengcheng and Neubig, Graham and Yih, Wen-tau and Riedel, Sebastian},
	year = {2020},
	pages = {8413--8426},
}

@article{levin_transfer_2023,
	title = {Transfer {Learning} with {Deep} {Tabular} {Models}},
	url = {https://openreview.net/forum?id=b0RuGUYo8pA},
	doi = {10.48550/arxiv.2206.15306},
	journal = {International Conference on Learning Representations},
	author = {Levin, Roman and Cherepanova, Valeriia and Schwarzschild, Avi and Bansal, Arpit and Bruss, C Bayan and Goldstein, Tom and Wilson, Andrew Gordon and Goldblum, Micah},
	year = {2023},
}

@article{pietruszka_stable_2022,
	title = {{STable}: {Table} {Generation} {Framework} for {Encoder}-{Decoder} {Models}},
	url = {https://app.readcube.com/library/12dc685d-b246-4fde-a409-59b857cce495/item/6bb1d68e-6a35-48c2-9e03-0779ac875f36},
	doi = {10.48550/arxiv.2206.04045},
	journal = {NeurIPS Table Representation Learning Workshop},
	author = {Pietruszka, Michał and Turski, Michał and Borchmann, Łukasz and Dwojak, Tomasz and Pałka, Gabriela and Szyndler, Karolina and Jurkiewicz, Dawid and Garncarek, Łukasz},
	year = {2022},
}

@article{somepalli_saint_2022,
	title = {{SAINT}: {Improved} {Neural} {Networks} for {Tabular} {Data} via {Row} {Attention} and {Contrastive} {Pre}-{Training}},
	url = {https://app.readcube.com/library/12dc685d-b246-4fde-a409-59b857cce495/item/c066849b-70f3-40d0-ae3e-2791b2c28c32},
	doi = {10.48550/arxiv.2106.01342},
	journal = {NeurIPS Table Representation Learning Workshop},
	author = {Somepalli, Gowthami and Goldblum, Micah and Schwarzschild, Avi and Bruss, C Bayan and Goldstein, Tom},
	year = {2022},
}

@article{loshchilov_decoupled_2017,
	title = {Decoupled {Weight} {Decay} {Regularization}},
	shorttitle = {{AdamW}},
	url = {https://app.readcube.com/library/12dc685d-b246-4fde-a409-59b857cce495/item/3ef196d5-88f7-4811-9f5a-5a9fba67ae50},
	doi = {10.48550/arxiv.1711.05101},
	journal = {International Conference on Learning Representations},
	author = {Loshchilov, Ilya and Hutter, Frank},
	year = {2017},
}

@article{OpenML2013,
author = {Vanschoren, Joaquin and van Rijn, Jan N. and Bischl, Bernd and Torgo, Luis},
title = {OpenML: Networked Science in Machine Learning},
journal = {SIGKDD Explorations},
volume = {15},
number = {2},
year = {2013},
pages = {49--60},
url = {http://doi.acm.org/10.1145/2641190.2641198},
doi = {10.1145/2641190.2641198},
publisher = {ACM},
address = {New York, NY, USA},
}

\clearpage
\appendix

\section{Additional Technical Details}
\label{app:technical-details}

\paragraph{Model configuration details}
As previously discussed, we investigate four different model configurations. 
The configurations are depicted in Table~\ref{tab:model-configs} with a detailed parameter count, including the tokenizer and projection layers in the case of single-table and cross-table pretraining, given in Table~\ref{tab:model-sizes-details}.

\paragraph{Hyperparameter details}
The full hyperparameter configuration of our approach and pretraining setup is given in Table~\ref{tab:hyperparameters} with model-specific ones detailed in Table~\ref{tab:training-configs}.
For the masking procedure, each value is masked randomly by drawing from the Binomial distribution of the corresponding masking fraction.

\begin{table}[h]
    \footnotesize
    \centering
    \caption{Investigated Transformer backbone model configurations. The XTab configuration is shown for comparison. The depicted parameters correspond solely to the backbone parameters and do not include the trainable parameters of the tokenizers or projection heads.}
    \begin{tabular}{lS[table-format=2.0]S[table-format=2.0]S[table-format=2.0]r@{\,}l}
  \toprule
  {Model} & {Embedding dimension} & {Number of heads} & {Number of layers} & \multicolumn{2}{c}{Parameter count} \\
  \midrule
  XTab~\cite{zhu_xtab_2023}  & 192   & 8   & 3 & \hspace{2.2em}740 & k\\ \arrayrulecolor{black!30}\midrule\arrayrulecolor{black}
  S     & 16    & 4   & 4 & 13  & k\\
  M     & 64    & 8   & 4 & 200 & k\\
  L     & 128   & 8   & 8 & 1.6 & M\\
  XL    & 192   & 16  & 36 & 16 & M\\
  \bottomrule
\end{tabular}
    \label{tab:model-configs}
\end{table}
\begin{table}[h]
    \footnotesize
    \centering
    \caption{Number of trainable parameters of the different investigated model configurations, including backbone, tokenizer, and projection head parameters. Note that, in the single-table case, the number of encoder and projection parameters depend on the datasets specifics, in particular the number of rows and categorical as well as numerical features.
Therefore, we state the minimum and maximum number of parameters across the five used benchmark datasets for reference.}
    \begin{tabular}{lrrrrr}
  \toprule
  \multirow{3}{*}{\vspace{-3mm}\textbf{Model}} & \multicolumn{5}{c}{\textbf{Parameter count}} \\ \cmidrule[1.25pt](lr){2-6} 
   & \multirow{3}{*}{\vspace{1mm}Backbone} & \multicolumn{2}{c}{Tokenizer} & \multicolumn{2}{c}{Projection Head}   \\\cmidrule(lr){3-4} \cmidrule(lr){5-6} 
   & & Single-table & Cross-table & Single-table & Cross-table\\
  \midrule
  S     & 13\,k    & 3\,k -- 10\,k    & - & 3\,k--11\,k & -   \\
  M     & 200\,k   & 11\,k -- 41\,k    & 2\,M & 11\,k--41\,k & 2\,M   \\
  L     & 1.6\,M   & 22\,k -- 82\,k    & 4\,M & 23\,k--83\,k & 4\,M   \\
  XL     & 16\,M   & -    & 6\,M & - & 6\,M   \\
  \bottomrule
\end{tabular}
    \label{tab:model-sizes-details}
\end{table}

\begin{table}[h]
    \footnotesize
	\begin{floatrow}
		\ffigbox[0.4\textwidth]
		{\captionof{table}{Detailed hyperparameter configuration of the proposed approach and pretraining setup.}%
		\label{tab:hyperparameters}}
		{\newcommand{\LINESEP}{0.5mm}
\begin{tabular}{lS[table-format=2]}
  \toprule
  {Hyperparameter} & {Value}\\
  \midrule
  Dropout rate     & 0.1    \\[\LINESEP]
  Masking rate     & 0.25   \\[\LINESEP]
  Optimizer     &    \\
  \quad Type     &  {AdamW}  \\
  \quad $\beta_0$     &  0.9  \\
  \quad$\beta_1$     &  0.999  \\
  \quad Weight decay &  0.01  \\
  \quad Learning rate init     &  0.00005  \\
  \quad Learning rate peak     &  0.001  \\
  \quad Learning rate final     &  0  \\[\LINESEP]
  Batch size     &  {256\,--\,2048}  \\[\LINESEP]
  Train samples     & {5\,M\,--\,75\,M}   \\[\LINESEP]
  Encoder     &    \\
  \quad Numerical encoder &  {Quantile enc.}  \\
  \quad Num. quantiles     &  25                 \\
  \quad Categorical encoder &  {Integer enc.}  \\
  \quad Categorical threshold &  20  \\[\LINESEP]
  Embedder     &    \\
  \quad Numerical embedder &  {Look-up emb.}  \\
  \quad Categorical embedder &  {Look-up emb.}  \\
  \bottomrule
\end{tabular}
\hfill}\hspace{3em}\hfill
		\ffigbox[0.5\textwidth]
		{\captionof{table}{Detailed model-specific batch size and training samples used with our approach. Note that the number of training samples corresponds to rows and not the total amount of tokens, which also on the numbers of columns per table.}%
		\label{tab:training-configs}}
		{\hfill\setlength\tabcolsep{1ex}
\begin{tabular}{lrrrr}
  \toprule
  \multirow{2}{*}{\vspace{-1mm}Model} & \multicolumn{2}{c}{Single-table} & \multicolumn{2}{c}{Cross-table} \\ \cmidrule(lr){2-3} \cmidrule(lr){4-5} 
   & {Batch size} & {Samples} & {Batch size} & {Samples}  \\
  \midrule
  S     & 2048   & 5\,M    & -   & -     \\
  M     & 2048   & 10\,M   & 512 & 75\,M \\
  L     & 2048   & 25\,M   & 512 & 75\,M \\
  XL    & -      & -       & 256 & 75\,M \\
  \bottomrule
\end{tabular}
}
	\end{floatrow}%
\end{table}
%
%
%

\clearpage
\section{Additional Results}
\label{app:additional-results}

\begin{figure}[h]
    \centering
    \begin{subfigure}[b]{0.475\textwidth}
        \includegraphics[width=\textwidth]{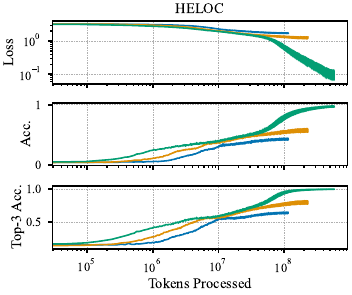}
    \end{subfigure}
    \hfill
    \begin{subfigure}[b]{0.475\textwidth}
        \includegraphics[width=\textwidth]{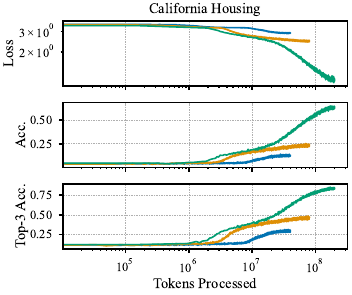}
    \end{subfigure}
    \medskip
    \begin{subfigure}[b]{0.475\textwidth}
        \includegraphics[width=\textwidth]{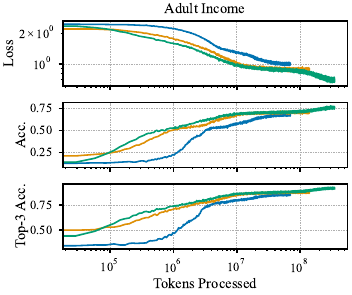}
    \end{subfigure}
    \hfill
    \begin{subfigure}[b]{0.475\textwidth}
        \includegraphics[width=\textwidth]{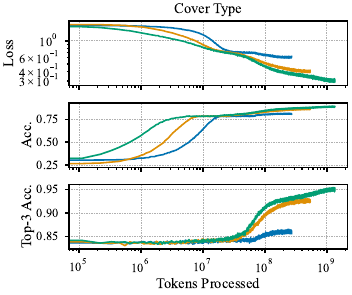}
    \end{subfigure}
    \medskip
    \begin{subfigure}[b]{0.475\textwidth}
        \includegraphics[width=\textwidth]{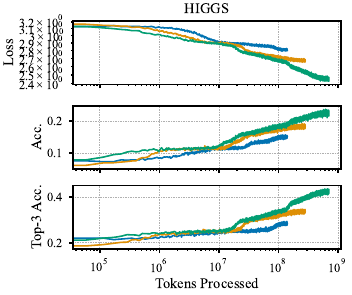}
    \end{subfigure}
    \hfill
    \begin{subfigure}[b]{0.475\textwidth}
        \includegraphics[width=\textwidth]{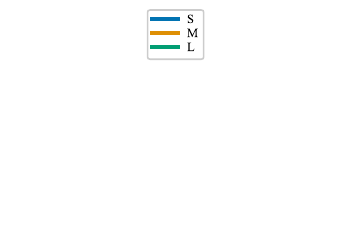}
    \end{subfigure}
    \caption{Training loss and imputation accuracy for the individual self-supervised single-table training using the five investigated benchmark datasets.}
    \label{fig:single-table-training-loss}
\end{figure}

\paragraph{Single-table pretraining}
The training loss and training imputation accuracy, including top-3 accuracy, are given in Figure~\ref{fig:single-table-training-loss}.
In general, we observe a strong imputation performance (on the training data), in particular for larger model sizes.
In fact, the results for the smaller datasets, namely HELOC and California Housing, show a potential of overfitting with the Large model configuration.
That is, we observe a double-descent-like training loss.
However, we do not observe an additional validation set as the remaining portions of the datasets are used for linear probing.
Hence, whether overfitting actually occurs or we are in an interpolation, i.e.\ grokking, regime, is speculative~\cite{belkin_fit_2021,nakkiran_deep_2021,power_grokking_2022}.

\begin{figure}[h]
    \centering
    \includegraphics[trim={8mm 3.55cm 0 0},clip]{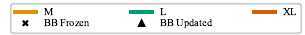}\\[2mm]
    \begin{subfigure}[b]{0.45\textwidth}
        \includegraphics[width=\textwidth]{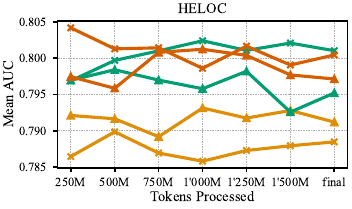}
    \end{subfigure}
    \hfill
    \begin{subfigure}[b]{0.45\textwidth}
        \includegraphics[width=\textwidth]{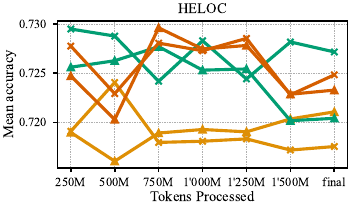}
    \end{subfigure}
    \medskip
    \begin{subfigure}[b]{0.45\textwidth}
        \includegraphics[width=\textwidth]{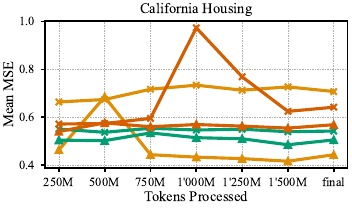}
    \end{subfigure}
    \hfill
    \begin{subfigure}[b]{0.45\textwidth}
        \includegraphics[width=\textwidth]{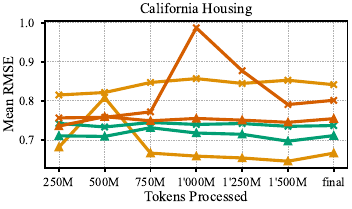}
    \end{subfigure}
    \medskip
    \begin{subfigure}[b]{0.45\textwidth}
        \includegraphics[width=\textwidth]{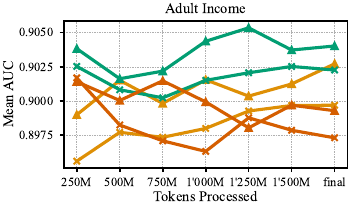}
    \end{subfigure}
    \hfill
    \begin{subfigure}[b]{0.45\textwidth}
        \includegraphics[width=\textwidth]{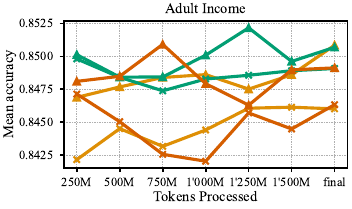}
    \end{subfigure}
    \medskip
    \begin{subfigure}[b]{0.45\textwidth}
        \includegraphics[width=\textwidth]{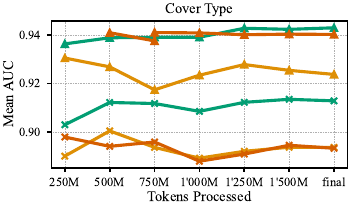}
    \end{subfigure}
    \hfill
    \begin{subfigure}[b]{0.45\textwidth}
        \includegraphics[width=\textwidth]{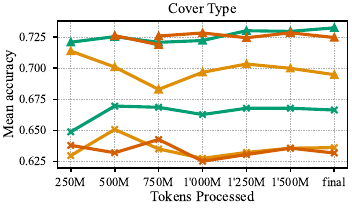}
    \end{subfigure}
    \medskip
    \begin{subfigure}[b]{0.45\textwidth}
        \includegraphics[width=\textwidth]{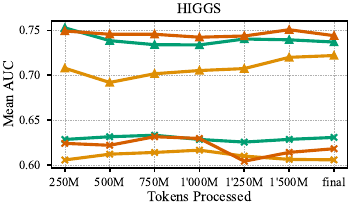}
    \end{subfigure}
    \hfill
    \begin{subfigure}[b]{0.45\textwidth}
        \includegraphics[width=\textwidth]{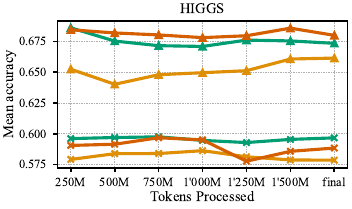}
    \end{subfigure}
    \caption{Linear probe evaluations using the pretrained model at different stages of pretraining for all considered benchmark datasets.}
    \label{fig:cross-table-checkpoints}
\end{figure}

\paragraph{Cross-table pretraining}
The linear probe accuracy in the cross-table pretraining case for different amounts of processed training tokens is depicted in Figure~\ref{fig:cross-table-checkpoints}.
That is, the cross-table pretrained models were checkpointed every 250\,M tokens.
All checkpointed models were then transferred to the individual benchmark dataset via self-supervised learning using the imputation loss in order to train a new tokenizer for each set, in full analogy to the evaluation of the final checkpoints depicted in the main text.
The obtained models are then evaluated via linear probing as previously discussed.
Again, we generally see that updating the backbone models during transfer on the specific downstream tasks is beneficial.
This is not to be confused with a supervised fine-tuning as we update the backbone model weights jointly with the encoder via the self-supervised imputation loss.
However, we observe, that regardless of the model size, the final downstream task performance does not significantly increase with longer pretraining.
This is somewhat in line with the observations made in the main paper, i.e.\, that the cross-table pretrained models seem to be bottlenecked in their performance, likely due to a lack of pretraining and transfer learning data.
This needs to be further investigated in future works.

\begin{figure}[h]
    \centering
    \begin{subfigure}[b]{0.475\textwidth}
        \includegraphics[width=\textwidth]{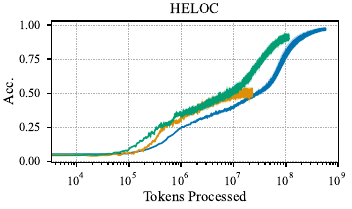}
    \end{subfigure}
    \hfill
    \begin{subfigure}[b]{0.475\textwidth}
        \includegraphics[width=\textwidth]{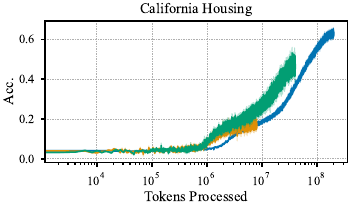}
    \end{subfigure}
    \medskip
    \begin{subfigure}[b]{0.475\textwidth}
        \includegraphics[width=\textwidth]{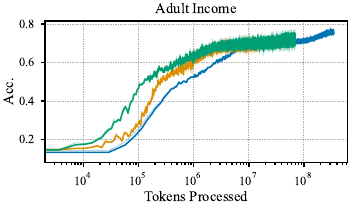}
    \end{subfigure}
    \hfill
    \begin{subfigure}[b]{0.475\textwidth}
        \includegraphics[width=\textwidth]{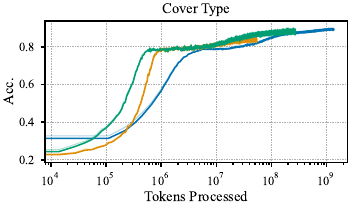}
    \end{subfigure}
    \medskip
    \begin{subfigure}[b]{0.475\textwidth}
        \includegraphics[width=\textwidth]{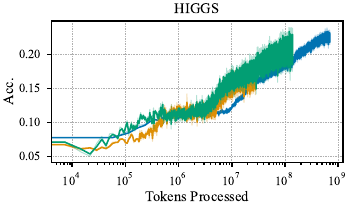}
    \end{subfigure}
    \hfill
    \begin{subfigure}[b]{0.475\textwidth}
        \includegraphics[width=\textwidth]{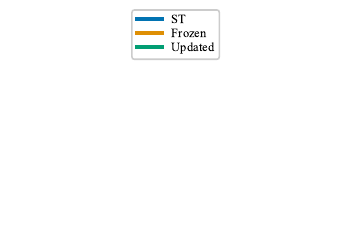}
    \end{subfigure}
    \caption{Imputation accuracy during the transfer learning of the cross-table pretrained models onto each benchmark dataset.}
    \label{fig:cross-table-transfer-loss}
\end{figure}

A similar observation can be made from the transfer learning loss and imputation accuracy curves as depicted in Figure~\ref{fig:cross-table-transfer-loss}.
We see that the imputation accuracy increases steeply with only few transfer learning steps, underlining the generalization capability of the models as an imputation system.
However, further (self-supervised) transfer learning does not further increase imputation accuracy.
Assuming that the imputation accuracy is an informative proxy for the downstream task performance, this would point towards a similar problem as previously discussed.

\clearpage
\section{Additional Dataset Details}
\label{app:dataset-details}

\paragraph{Benchmark datasets}
Detailed properties of the used benchmark datasets HELOC, California Housing, Adult Income, Cover Type, and HIGGS are stated in Table~\ref{tab:benchmark-datasets}.

\begin{figure}[h]
    \centering
    \includegraphics{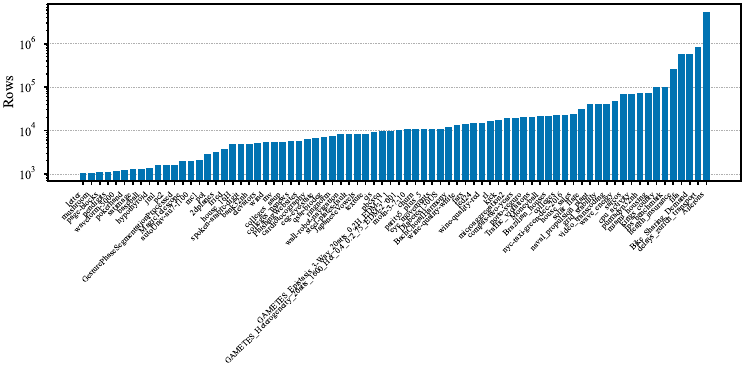}
    \hfill
    \includegraphics{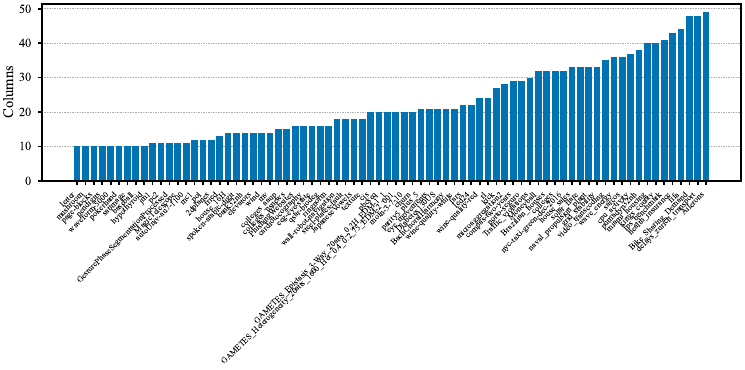}
    \caption{Row and column statistics of the datasets used in our curated pretraining corpus.}
    \label{fig:dataset-rows-cols}
\end{figure}

\paragraph{Pretraining corpus}
As previously described, we collect a total of 74 datasets from various sources as our pretraining corpus.
The column and row statistics of the used datasets are shown in Figure~\ref{fig:dataset-rows-cols}.
Further details of all used datasets are given in Table~\ref{tab:training-datasets}.

\begin{table}[h]
    \footnotesize
    \centering
    \caption{Properties of the used benchmark datasets. Throughout, evaluation is performed using a 5-fold cross-validation on the eval split.}
    \sisetup{table-number-alignment = right,table-text-alignment = right}
\begin{tabular}{lS[table-format=2]S[table-format=2.0]S[table-format=2.0]S[table-format=2.0]r@{\,}l}
    \toprule
    Task                    & {HELOC}   & {California Housing} & {Adult Income} & {Cover Type} & \hspace{0.7em}HIGGS \\
    \midrule
    Task type               & {binary}  & {regression}         & {binary}       & {multiclass} & {binary} & \\
    Samples                 & 7519      & 16512                & 39073          & 464809       & \num{8800000} & \\
    Features (all)          & 24        & 9                    & 15             & 55           & 29 & \\
    Features (numerical)\hspace{-2em}    & 16        & 8                    & 1              & 10           & 24  & \\
    Missing values          & 0         & 0                    & 5259           & 0            & 0& \\
    Pretrain split          & 60\,\%         & 60\,\%        & 60\,\%          & 60\,\%            & 60 &\% \\
    Eval split          & 40\,\%         & 40\,\%        & 40\,\%          & 40\,\%            & 40 & \% \\
    \bottomrule
\end{tabular}
    \label{tab:benchmark-datasets}
\end{table}

\begin{landscape}
\begin{table}
        \footnotesize
        \centering
        \caption{Details of the datasets used in our curated pretraining corpus.}
        \begin{tabularx}{\linewidth}{XXlS[table-format=6]S[table-format=7]S[table-format=2]S[table-format=2]S[table-format=2]S[table-format=2]S[table-format=3]S[table-format=2.1]}
\toprule
                                              {Name} &                               {Collection} &       {Task} &  {Classes} &  {Rows} &  {Columns} &  {Numer.\ features} &  {Cat.\ features} &  {Unique cat.} &  {NANs in \%} \\
\midrule
                                wine-quality-white &                automl\_benchmark training & multiclass &             7 &       4898 &         11 &                  11 &                     0 &                  0 &       0.0 \\
                                  wine-quality-red &                automl\_benchmark training & multiclass &             6 &       1599 &         11 &                  11 &                     0 &                  0 &       0.0 \\
                                              wind &                automl\_benchmark training &     binary &             2 &       6574 &         14 &                  14 &                     0 &                  0 &       0.0 \\
                                     waveform-5000 &                automl\_benchmark training & multiclass &             3 &       5000 &         40 &                  40 &                     0 &                  0 &       0.0 \\
                                       wave\_energy &                             openml\_ctr23 & regression &         71993 &      72000 &         48 &                  48 &                     0 &                  0 &       0.0 \\
                             wall-robot-navigation &                automl\_benchmark training & multiclass &             4 &       5456 &         24 &                  24 &                     0 &                  0 &       0.0 \\
                                 video\_transcoding &                             openml\_ctr23 & regression &         10960 &      68784 &         18 &                  16 &                     2 &                  8 &       0.0 \\
                                           texture &                automl\_benchmark training & multiclass &            11 &       5500 &         40 &                  40 &                     0 &                  0 &       0.0 \\
                                steel-plates-fault &                automl\_benchmark training &     binary &             2 &       1941 &         33 &                  33 &                     0 &                  0 &       0.0 \\
                               spoken-arabic-digit &                automl\_benchmark training & multiclass &            10 &     263256 &         14 &                  14 &                     0 &                  0 &       0.0 \\
                                       solar\_flare &                             openml\_ctr23 & multiclass &             8 &       1066 &         10 &                   2 &                     8 &                 27 &       0.0 \\
                                          satimage &                automl\_benchmark training & multiclass &             6 &       6430 &         36 &                  36 &                     0 &                  0 &       0.0 \\
                                            sarcos &                             openml\_ctr23 & regression &         11414 &      48933 &         21 &                  21 &                     0 &                  0 &       0.0 \\
                                                rl &                automl\_benchmark training &     binary &             2 &      31406 &         22 &                   8 &                    14 &                 89 &       4.3 \\
                                          ringnorm &                automl\_benchmark training &     binary &             2 &       7400 &         20 &                  20 &                     0 &                  0 &       0.0 \\
                                       qsar-biodeg &                automl\_benchmark training &     binary &             2 &       1055 &         41 &                  41 &                     0 &                  0 &       0.0 \\
                                       pumadyn32nh &                             openml\_ctr23 & regression &          8191 &       8192 &         32 &                  32 &                     0 &                  0 &       0.0 \\
                                      porto-seguro &                automl\_benchmark training &     binary &             2 &     595212 &         37 &                  12 &                    25 &                102 &       3.8 \\
                                               pol &                automl\_benchmark training &     binary &             2 &      15000 &         48 &                  48 &                     0 &                  0 &       0.0 \\
                                         pokerhand &                automl\_benchmark training & multiclass &            10 &     829201 &         10 &                   5 &                     5 &                 20 &       0.0 \\
                                         pendigits &                automl\_benchmark training & multiclass &            10 &      10992 &         16 &                  16 &                     0 &                  0 &       0.0 \\
                                               pc2 &                automl\_benchmark training &     binary &             2 &       5589 &         36 &                  36 &                     0 &                  0 &       0.0 \\
                                            pbcseq &                automl\_benchmark training &     binary &             2 &       1945 &         18 &                  12 &                     6 &                 13 &       3.2 \\
                                    parity5\_plus\_5 &                automl\_benchmark training &     binary &             2 &       1124 &         10 &                   0 &                    10 &                 20 &       0.0 \\
                                       page-blocks &                automl\_benchmark training & multiclass &             5 &       5473 &         10 &                  10 &                     0 &                  0 &       0.0 \\
                           nyc-taxi-green-dec-2016 &              automl\_benchmark regression & regression &          1811 &     581835 &         18 &                   9 &                     9 &                 22 &       0.0 \\
                            naval\_propulsion\_plant &                             openml\_ctr23 & regression &            51 &      11934 &         14 &                  14 &                     0 &                  0 &       0.0 \\
                                                mv &                automl\_benchmark training &     binary &             2 &      40768 &         10 &                   7 &                     3 &                  7 &       0.0 \\
                                          mushroom &                automl\_benchmark training &     binary &             2 &       8124 &         22 &                   0 &                    22 &                117 &       1.4 \\
                                       mofn-3-7-10 &                automl\_benchmark training &     binary &             2 &       1324 &         10 &                   0 &                    10 &                 20 &       0.0 \\
                                 microaggregation2 &                automl\_benchmark training & multiclass &             5 &      20000 &         20 &                  20 &                     0 &                  0 &       0.0 \\
                                     miami\_housing &                             openml\_ctr23 & regression &          2111 &      13932 &         15 &                  15 &                     0 &                  0 &       0.0 \\
                                               mc1 &                automl\_benchmark training &     binary &             2 &       9466 &         38 &                  38 &                     0 &                  0 &       0.0 \\
                                            letter &                automl\_benchmark training & multiclass &            26 &      20000 &         16 &                  16 &                     0 &                  0 &       0.0 \\
                                             led24 &                automl\_benchmark training & multiclass &            10 &       3200 &         24 &                   0 &                    24 &                 48 &       0.0 \\
                                      kings\_county &                             openml\_ctr23 & regression &          4028 &      21613 &         21 &                  17 &                     4 &                 45 &       0.0 \\
                                              kick &                automl\_benchmark training &     binary &             2 &      72983 &         32 &                  14 &                    18 &                134 &       6.4 \\
                                               jm1 &                automl\_benchmark training &     binary &             2 &      10885 &         21 &                  21 &                     0 &                  0 &       0.0 \\
\bottomrule
\end{tabularx}

        \label{tab:training-datasets}
    \end{table}
\end{landscape}
\begin{landscape}
\begin{table}
        \footnotesize
        \centering
        \begin{tabularx}{\linewidth}{XXlS[table-format=6]S[table-format=7]S[table-format=2]S[table-format=2]S[table-format=2]S[table-format=2]S[table-format=3]S[table-format=2.1]}
\toprule
                                              {Name} &                               {Collection} &       {Task} &  {Classes} &  {Rows} &  {Columns} &  {Numer.\ features} &  {Cat.\ features} &  {Unique cat.} &  {NANs in \%} \\
\midrule
                                       hypothyroid &                automl\_benchmark\_training & multiclass &             4 &       3772 &         29 &                   7 &                    22 &                 47 &       5.5 \\
                                       house\_sales &              automl\_benchmark\_regression & regression &          4028 &      21613 &         21 &                  20 &                     1 &                  0 &       0.0 \\
                                         house\_16H &                automl\_benchmark\_training &     binary &             2 &      22784 &         16 &                  16 &                     0 &                  0 &       0.0 \\
                                  health\_insurance &                             openml\_ctr23 & regression &            75 &      22272 &         11 &                   4 &                     7 &                 21 &       0.0 \\
                                    grid\_stability &                             openml\_ctr23 & regression &         10000 &      10000 &         12 &                  12 &                     0 &                  0 &       0.0 \\
                                             fried &                automl\_benchmark\_training &     binary &             2 &      40768 &         10 &                  10 &                     0 &                  0 &       0.0 \\
                                     fps\_benchmark\_&                             openml\_ctr23 & regression &          2675 &      24624 &         43 &                  30 &                    13 &                 96 &       6.6 \\
                                              fifa &                             openml\_ctr23 & regression &           133 &      19178 &         28 &                  27 &                     1 &                  0 &       0.0 \\
                                              fars &                automl\_benchmark\_training & multiclass &             8 &     100968 &         29 &                  14 &                    15 &                 93 &       0.0 \\
                                     eye\_movements &                automl\_benchmark\_training & multiclass &             3 &      10936 &         27 &                  24 &                     3 &                  6 &       0.0 \\
                                         elevators &                automl\_benchmark\_training &     binary &             2 &      16599 &         18 &                  18 &                     0 &                  0 &       0.0 \\
                                     eeg-eye-state &                automl\_benchmark\_training &     binary &             2 &      14980 &         14 &                  14 &                     0 &                  0 &       0.0 \\
                           delays\_zurich\_transport & tabular\_benchmark\_ categorical\_regression & regression &          4082 &    5465575 &         11 &                   8 &                     3 &                 12 &       0.0 \\
                                      cpu\_activity &                             openml\_ctr23 & regression &            56 &       8192 &         21 &                  21 &                     0 &                  0 &       0.0 \\
                                  compas-two-years &                automl\_benchmark\_training &     binary &             2 &       5278 &         13 &                   7 &                     6 &                 12 &       0.0 \\
                                   colleges\_usnews &                automl\_benchmark\_training &     binary &             2 &       1302 &         33 &                  32 &                     1 &                 51 &      18.2 \\
                                     colleges\_aaup &                automl\_benchmark\_training &     binary &             2 &       1161 &         15 &                  13 &                     2 &                 56 &       1.5 \\
                                          colleges &              automl\_benchmark\_regression & regression &          4502 &       7063 &         44 &                  32 &                    12 &                206 &      33.5 \\
                                               cjs &                automl\_benchmark\_training & multiclass &             6 &       2796 &         33 &                  31 &                     2 &                 68 &      73.8 \\
                                             churn &                automl\_benchmark\_training &     binary &             2 &       5000 &         20 &                  16 &                     4 &                 17 &       0.0 \\
                                  cardiotocography &                automl\_benchmark\_training & multiclass &            10 &       2126 &         35 &                  35 &                     0 &                  0 &       0.0 \\
                                          baseball &                automl\_benchmark\_training & multiclass &             3 &       1340 &         16 &                  15 &                     1 &                  7 &       0.1 \\
                                          bank32nh &                automl\_benchmark\_training &     binary &             2 &       8192 &         32 &                  32 &                     0 &                  0 &       0.0 \\
                                 autoUniv-au7-1100 &                automl\_benchmark\_training & multiclass &             5 &       1100 &         12 &                   8 &                     4 &                 10 &       0.0 \\
                                Traffic\_violations &                automl\_benchmark\_training & multiclass &             3 &      70340 &         20 &                   1 &                    19 &                268 &       0.2 \\
                                  PhishingWebsites &                automl\_benchmark\_training &     binary &             2 &      11055 &         30 &                   0 &                    30 &                 68 &       0.0 \\
                                         Moneyball &              automl\_benchmark\_regression & regression &           374 &       1232 &         14 &                   8 &                     6 &                 66 &      20.9 \\
                                    MagicTelescope &                automl\_benchmark\_training &     binary &             2 &      19020 &         10 &                  10 &                     0 &                  0 &       0.0 \\
                                    JapaneseVowels &                automl\_benchmark\_training & multiclass &             9 &       9961 &         14 &                  14 &                     0 &                  0 &       0.0 \\
                 Gesture Phase Segmentation Processed &                automl\_benchmark\_training & multiclass &             5 &       9873 &         32 &                  32 &                     0 &                  0 &       0.0 \\
GAMETES Heterogeneity 20atts\ 600\_Het\_0.4\_0.2\_7 &                automl\_benchmark\_training &     binary &             2 &       1600 &         20 &                   0 &                    20 &                 59 &       0.0 \\
       GAMETES Epistasis 3-Way 20atts 0.2H EDM-1\_1 &                automl\_benchmark\_training &     binary &             2 &       1600 &         20 &                   0 &                    20 &                 60 &       0.0 \\
                                     Diabetes130US &                automl\_benchmark\_training & multiclass &             3 &     101766 &         49 &                  13 &                    36 &                130 &       0.0 \\
                                  Brazilian\_houses &              automl\_benchmark\_regression & regression &          5751 &      10692 &         12 &                   8 &                     4 &                 44 &       0.0 \\
                               Bike\_Sharing\_Demand & tabular\_benchmark\_ categorical\_regression & regression &           869 &      17379 &         11 &                   6 &                     5 &                 14 &       0.0 \\
                                 BachChoralHarmony &                automl\_benchmark\_training & regression &           102 &       5665 &         16 &                   2 &                    14 &                102 &       0.0 \\
                                          Ailerons &   tabular\_benchmark\_ numerical\_regression & multiclass &            35 &      13750 &         33 &                  33 &                     0 &                  0 &       0.0 \\
                                          2dplanes &                automl\_benchmark\_training &     binary &             2 &      40768 &         10 &                  10 &                     0 &                  0 &       0.0 \\
\bottomrule
\end{tabularx}
    \end{table}
\end{landscape}

\end{document}